# Gated Feedback Refinement Network for Coarse-to-Fine Dense Semantic Image Labeling

Md Amirul Islam, Mrigank Rochan, Shujon Naha, Neil D. B. Bruce, and Yang Wang

*Abstract*—Effective integration of local and global contextual information is crucial for semantic segmentation and dense image labeling. We develop two encoder-decoder based deep learning architectures to address this problem. We first propose a network architecture called *Label Refinement Network* (**LRN**) that predicts segmentation labels in a coarse-to-fine fashion at several spatial resolutions. In this network we also define loss functions at several stages to provide supervision at different stages of training. However, there are limits to the quality of refinement possible if ambiguous information is passed forward. In order to address this issue we also propose *Gated Feedback Refinement Network* (**G-FRNet**) that addresses this limitation. Initially, G-FRNet makes a coarse-grained prediction which it progressively refines to recover details by effectively integrating local and global contextual information during the refinement stages. This is achieved by gate units proposed in this work, that control information passed forward in order to resolve ambiguity. Experiments were conducted on four challenging dense labeling datasets (CamVid, PASCAL VOC 2012, Horse-Cow Parsing, PASCAL-Person-Part, and SUN-RGBD). G-FRNet achieves state-of-the-art semantic segmentation results on the CamVid and Horse-Cow Parsing datasets, and produces results competitive with the best performing approaches that appear in the literature for the other three datasets.

*Index Terms*—Semantic Segmentation, Encoder-Decoder Network, Coarse-to-Fine, Gating Mechanism, Deep Supervision.

## I. INTRODUCTION

In recent years, there have been significant advances in deep learning applied to problems in computer vision. This has been met with a great deal of success, and has given rise to proliferation of significant variety in the structure of neural networks. Many current deep learning models [1], [2], [3] apply a cascade comprised of repeated convolutional stages, followed by spatial pooling. Down-sampling by pooling allows for a very large pool of distinct and rich features, albeit at the expense of spatial precision. For recognition problems, the loss of spatial precision is not especially problematic. However, dense image labeling problems (e.g. semantic segmentation [4]) require pixel-level precision. For example, consider Fig. 1 that has fine objects such as a column-pole, pedestrian or bicyclist which require extraction of very fine details in order to be segmented accurately. Naturally, ConvNets try to extract exactly this type of fine-grained high spatial-frequency variation in the early convolution stages, even though corresponding object specific representations may not emerge until very late in the

M. A. Islam, M. Rochan, N. Bruce, and Y. Wang are with the Department of Computer Science, University of Manitoba, Winnipeg, MB R3T 2N2
E-mail: {amirul, mrochan, bruce, ywang}@cs.umanitoba.ca
S. Naha is with the School of Informatics and Computing, Indiana University, Bloomington, IN. E-mail: snaha@iu.edu

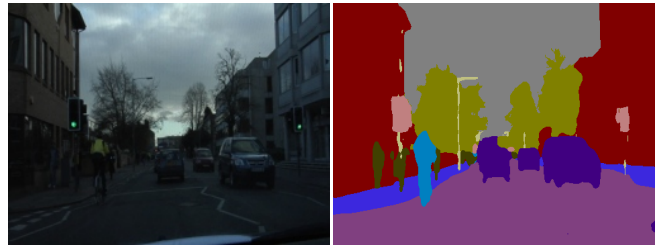

Fig. 1. Labeling objects such as the column-pole, pedestrian or bicyclist shown above requires low-level finer details as well as high-level contextual information, despite of varying light scenes. In this paper, we propose a gated feedback refinement network, which can be used with any bottom-up, feed-forward ConvNet. We show that the finer features learnt by our approach lead to significantly improved semantic segmentation.

feed-forward ConvNet architecture. A question that naturally follows from this is: How can we incorporate these fine details in the segmentation process to get precise labeling while also carrying a rich feature-level representation?

Deep learning models for dense image labeling problems typically involve a decoding process that gradually recovers a pixel level specification of categories. In some cases this *decoding* is done in one step [4], while in other instances, both the encoding of patterns, and gradual recovery of spatial resolution are hierarchical. It is interesting to note that this mirrors the observed computational structure of human vision wherein space is abstracted away in favour of rich features, and recognition of patterns precedes their precise localization [5].

Some models that have shown success for segmentation problems [6], [7] share a common structure involving stage-wise encoding of an input image, followed by stage-wise decoding to recover a per-pixel categorization. At an abstract level, this is reminiscent of a single network that involves a feed-forward pass, followed by a recurrent pass from the top layer downward where additional computation and refinement ensues. There are tangible distinctions though, in that decoding is typically driven only by information flow that satisfies solving a specific labeling problem, and that all decoding may be informed only by the representation carried by the highest encoder layer.

At the deepest stage of encoding, one has the richest possible feature representation, and relatively poor spatial resolution from a per-neuron perspective. While spatial resolution may be poor from a per-neuron perspective, this does not necessarily imply that recovery of precise spatial information is impossible. For example, a coarse coding strategy [8], [9] may allow for a high degree of precision in spatial localization but at the expense of the diversity of features encoded and involved in discrimination. An important implication of this, is that provided the highest layer does not require the power to precisely localize patterns, a much richer feature level representation is possible.

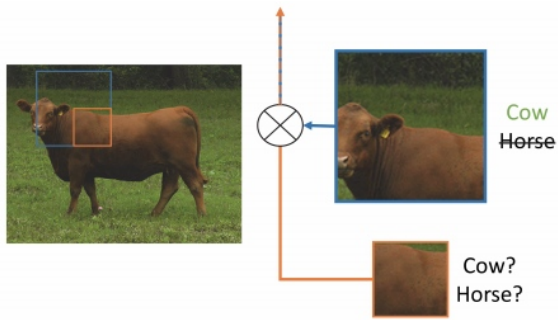

Fig. 2. An illustration of the relationship between receptive field size across layers, and ambiguity that may arise. In this case, the larger (and more discriminative) receptive field (blue) resides at a deeper layer of the network, and may be of value in refining the representation carried by an earlier layer (orange) to resolve ambiguity and improve upon labeling performance.

Information carried among earlier layers of encoding do have greater spatial locality, but may be less discriminative. Given that there is an extant representation of image characteristics at every layer, it is natural to assume that value may be had in leveraging earlier encoding representations at the decoding stage. In this manner, spatial precision that may be lost at deep layers in encoding may be gradually recovered from earlier representations. This removes some of the onus on deeper layers to represent highly discriminative characteristics of the image, while simultaneously facilitating precise localization. This intuition appears in the model we propose, as seen in connections between encoder layers and decoder layers in our network. This implies the shift in responsibility among encoding layers, and the associated discriminative power or capacity deeper in the network.

If one were to label categories within the image using only early layers, this may be problematic, especially in instances where local parts are ambiguous. The re-use of information from earlier encoder layers at the decoding stage is weakened by their lack of discrimination. For example, if one assumes reliance on convolution, and unpooling (which involves a fixed set of weights) to recover information and ultimately assign labels, this implies that any ambiguous representations are necessarily involved in decoding, which may degrade the quality of predictions. For example, while a convolutional layer deep within the network may provide strong discrimination between a cow and a horse, representations from earlier layers may be specific to animals, but express confidence for both. If this confidence is passed on to the decoding stage, and a fixed scheme for combining these representations is present, this contributes to error in labeling. This observation forms the motivation for the most novel and important aspect of our proposed model and this intuition is illustrated in Fig. 2. While information from early encoding layers may be of significant value to localization, it is sensible to filter this information such that categorical ambiguity is reduced. Moreover, it is natural to use deeper, more discriminative layers in filtering information passed on from less discriminative, but more finely localized earlier layers.

The precise scheme that achieves this is discussed in detail in the remainder of this paper. We demonstrate that a high degree of success may be achieved across a variety of benchmarks, using a relatively simple model structure in applying a canonical gating mechanism that may be applied to any network comprised of encoder and decoder components. This is also an area in which parallels may be drawn to neural information processing in humans, wherein more precisely localized representations that may be ambiguous are modulated or gated by higher-level features, iteratively and in a top-down fashion [10].

Based on the above observations, it is evident that features from all levels of a neural hierarchy (high+low) are of value, or even necessary for accurate semantic segmentation. High-level or global features usually help in recognizing the category label while low-level or local features can help in assigning precise boundaries to the objects since spatial details tend to get lost in high-level/global features. In this paper, we propose a novel end-to-end network architecture that effectively exploits multi-level features in semantic segmentation. We summarize our main contributions as follows:

- We introduce a new perspective on semantic segmentation, or more generally, pixel-wise labeling of images. Instead of predicting the final segmentation result in a single shot, we propose to solve the problem in a coarse-to-fine fashion by first predicting a coarse labeling, then progressively refining this coarse grained prediction towards finer scale results.
- We introduce a novel gating mechanism to modulate how information is passed from the encoder to the decoder in the network. The gating mechanism allows the network to filter out ambiguity concerning object categories as information is passed through the network. The proposed approach is the first that uses a gating mechanism in an encoder-decoder framework for the task of semantic segmentation in order to combine local and global contextual information.
- Unlike most of the previous methods that only have supervision at the end of their network, our model has supervision at multiple resolutions in the network. Although we focus on semantic image segmentation in this paper, our network architecture is general enough to be used for any pixel-wise labeling task.

An earlier version of this work (G-FRNet) appeared in [11]. We extend G-FRNet in several ways. First, we present Label Refinement Network (LRN), a coarse-to-fine refinement network that inspired G-FRNet. Then, we conduct experiments on two more datasets (Pascal-Person-Part and SUN-RGBD) in addition to the datasets used in G-FRNet. Additionally, we examine the role of deep supervision technique in the proposed network on Pascal VOC 2012 dataset. Finally, we also explore alternate design choices for gating mechanism that includes additive interaction instead of multiplicative interaction.

## II. RELATED WORK

CNNs have shown tremendous success in various visual recognition tasks, e.g. image classification [13], object detection [14] and action recognition [15]. Recent approaches have also shown enhanced discriminative power of features within CNNs by increasing the depth of the network [1], [16], [2]. Dense labeling tasks, such as semantic segmentation, have also benefited from such deep networks. Recently, there has been work on adapting CNNs for pixel-wise image labeling problems such as semantic segmentation [6], [17], [18], [19], [4], [20], [7], [21].

Long et al. [4] proposed the first semantic segmentation network that trained fully convolutional networks (FCN) in an end-to-end fashion to accomplish pixel-wise prediction corresponding to a whole image. Their network is based on VGG-16 [1]. They define a novel skip architecture that combines semantic information with deep, coarse, and appearance based information.

F-CNN based approaches have one major limitation in that they produce a segmentation map of relatively low spatial resolution. There exist a number of methods that address this limitation by generating segmentation maps of higher spatial resolution. Recent methods including DeepLab-CRF [17] and DeepLabv2 [22] predict a mid-resolution label map by controlling the resolution of feature responses within the network, then directly upsampling to the original spatial resolution of the image by bi-linear interpolation. Finally a dense CRF is applied on top of the final prediction to refine boundaries. CRF-RNN [23] extended the DeepLab [17] network to include an end-to-end learning of the CNN and dense CRF. Recently, Yu et al. [24] introduced dilated convolution to expand the effective receptive field of feature maps to encode extra contextual information within local features that brought significant benefit to the ConvNet architectures in terms of performance.

Directly related to our work is the idea of multi-scale processing in computer vision. Early work on Laplacian pyramids [25] is a classic example of capturing image structures at multiple scales. Eigen et al. [26] use a multi-scale CNN for predicting depths, surface normals and semantic labels from a single image. Denton et al. [27] use a coarse-to-fine strategy in the context of image generation. Honari et al. [28] combine coarse-to-fine features in CNNs for facial keypoint localization. The Hypercolumn [19] method leverages features from intermediate layers to generate final predictions via stage-wise training. A recent pixel-wise labeling architecture named PixelNet [29] follows the hypercolumn strategy to combine contextual information with few predictions. SegNet [6] and DeconvNet [7] apply skip connections in the form of deconvolution layers to exploit the features produced in encoder stage to refine predictions at the time of decoding.

Few existing methods have used the idea of defining loss functions at different stages in the network to provide additional supervision in learning, and this is a characteristic of the models proposed in this paper. The Inception model [16] uses auxiliary classifiers at the lower stages of the network to encourage the network to learn discriminative features. Lee et al. [30] propose a deeply-supervised network for image classification. Xie et al. [31] use a similar idea for edge detection.

Architecturally, our work is closest to the strategies proposed in a few existing works [32], [33], [34], [35], [36] in which coarse-grained prediction maps are refined to generate final predictions by top-down modulation. However, how to effectively integrate fine grained simplistic features with coarse-level complex features still remains a open question. Pinheiro et al. [34] proposed a ConvNet architecture that has refinement modules with top-down modulation and skip connections to refine segmentation proposals. Ranjan et al. [36] also propose a spatial pyramid network that initially predicts a low-resolution optical flow map and iteratively uses the lower layer information to obtain high-resolution optical flow.

There is strong evidence of top-down modulation and feedback refinement in neural information processing within human and primate visual pathways [37], [38], [39], [10], wherein more precisely localized representations that may be ambiguous are modulated or gated by higher-level features, as well as act as attentional mechanism for selecting unambiguous features. Our proposed model is based on this intuition and integrates this style of information processing scheme within ConvNets in a fully encapsulated end-to-end trainable model.

## III. BACKGROUND

In this section, we describe background most relevant to our proposed model.

**Encoder-Decoder Architecture:** Our model (Fig. 4) is based on a deep encoder-decoder architecture (e.g. [6], [7]) used for dense image labeling problems such as semantic segmentation. The encoder network extracts features from an image and the decoder network produces semantic segmentation from the features generated by the encoder network. The encoder network is typically a CNN with alternating layers of convolution, pooling, non-linear activation, etc. The output of each convolution layer in the encoder network can be interpreted as features with different receptive fields. Due to spatial pooling, the spatial dimensions of the feature map produced by the encoder network are smaller than the original image. The subsampling layers in CNNs allow the networks to extract high-level features that are translation invariant, which are crucial for image classification. However, they reduce the spatial extent of the feature map at each layer in the network. Let $I \in \mathbb{R}^{h \times w \times d}$ be the input image (where $h$, $w$ are spatial dimensions $d$ is the color channel dimension) and $f(I) \in \mathbb{R}^{h' \times w' \times d'}$ be the feature map volume at the end of the encoder network. The feature map $f(I)$ has smaller spatial dimensions than the original image, i.e. $h' < h$ and $w' < w$. In order to produce full-sized segmentation results as the output, an associated decoder network is applied to the output of the encoder network to produce output that matches the spatial size of the original image. The fundamental differences between our work and various research contributions in prior work mainly lie in choices of the structure and composition of the decoder network. The decoder in SegNet [6] progressively enlarges the feature map using an upsampling technique without learnable parameters. The decoder in DeconvNet [7] is similar, but has learnable parameters. FCN [4] uses a single layer interpolation for deconvolution. The decoder network in general enlarges the feature map using upsampling and unpooling in order to produce the final semantic segmentation result. Many popular CNN-based semantic segmentation models fall into this encoder-decoder framework, e.g. FCN [4], SegNet [6], DeconvNet [7].

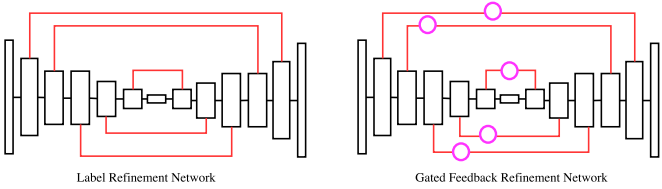

Fig. 3. Overview of our Label Refinement network and Gated Feedback Refinement Network (G-FRNet). In LRN, each of the decoder layers has long range connections to its corresponding encoder layer whereas G-FRNet adds a gate between the long range connections to modulate the influence of these connections.

**Skip Connections:** In a standard encoder-decoder architecture, the feature map from the top layer of the encoder network is used as the input for the decoder network. This feature map contains high-level features that tend to be invariant to "nuisance factors" such as small translation, illumination, etc. This invariance is crucial for certain high-level tasks such as object recognition, but is not ideal for many dense image labeling tasks (e.g. semantic segmentation) that require precise pixel-wise information, since important relationships may be abstracted away. One possible solution is to use "skip connections" [19], [4]. A skip connection directly links an encoder layer to a decoder layer. Since the bottom layers in the encoder network tend to contain precise pixel-wise information, the skip connections allow this information to be directly passed to the decoder network to produce the final segmentation result.

In the following sections, we firstly describe our base network called Label Refinement Network (LRN) (Sec. IV). Then we discuss our final network Gated Feedback Refinement Network (G-FRNet) (Sec. V) which builds on LRN. Fig. 3 illustrates the basic difference between these two network architectures.

## IV. Label Refinement Network

In this section, we describe a novel network architecture called the *Label Refinement Network (LRN)* for semantic segmentation. The architecture of LRN is the simpler version of the one shown in Fig. 4, where the difference is that there are no Gate Units in LRN. Similar to prior work [6], [4], [7], LRN also uses an encoder-decoder framework. The encoder network of LRN is similar to that of SegNet [6] which is based on the VGG16 network [1]. The novelty of LRN lies in the decoder network. Instead of making the prediction at the end of the network, the decoder network in LRN makes predictions in a coarse-to-fine fashion at several stages. In addition, LRN also has deep supervision which is aimed to provide supervision early in the network by adding loss functions at multiple stages (not just at the last layer) of the decoder network.

The LRN architecture is motivated by the following observations: Due to the convolution and subsampling operations, the feature map $f_7$ obtained at the end of the encoder network mainly contains high-level information about the image (e.g. objects). Spatially precise information is lost in the encoder network, and therefore $f_7$ cannot be used directly to recover a full-sized semantic segmentation which requires pixel-precision information. However, $f_7$ contains enough information to produce a *coarse* semantic segmentation. In particular, we can use $f_7$ to produce a segmentation map $P_m^G$ of spatial dimensions $h' \times w'$, which is smaller than the original image dimensions $h \times w$. Our decoder network then progressively refines the segmentation map $P_m^G$. Note that one can interpret $P_m^G$ as the feature map in the decoder network in most work (e.g. [6], [4], [7]). However, our model enforces the channel dimension of $P_m^G$ to be the same as the number of class labels, so $P_m^G$ can be considered as a (soft) label map.

The network architecture in Fig. 4 has 7 convolution layers in the encoder. We use $f_k(I) \in \mathbb{R}^{h_k \times w_k \times d_k}$ $(k = 1, 2, ..., 7)$ to denote the feature map after the $k$-th convolution layer. After the last convolution layer of the encoder network, we use a $3 \times 3$ convolution layer to convert the convolution feature map $f_7(I) \in \mathbb{R}^{h_7 \times w_7 \times C}$ to $P_m^G \in \mathbb{R}^{h_7 \times w_7 \times C}$, where $C$ is the number of class labels. We then define a loss function on $P_m^G$ as follows. Let $Y \in \mathbb{R}^{h \times w \times C}$ be the ground-truth segmentation map, where the label on each pixel is represented as a $C$-dimensional vector using the one-shot representation. We use $R_1(Y) \in \mathbb{R}^{h_7 \times w_7 \times C_7}$ to denote the segmentation map obtained by resizing $Y$ to have the same spatial dimensions as $P_m^G$. We can then define a loss function $\ell_1$ (cross entropy loss is used) to measure the difference between the resized ground-truth $R_1(Y)$ and the coarse prediction $P_m^G$ (after softmax). In other words, these operations can be written as:

$$P_m^G = \text{conv}_{3 \times 3}\big(f_7(I)\big)$$
$$\ell_1 = \text{Loss}\Big(R_1(Y), \text{softmax}\big(P_m^G\big)\Big) \qquad (1)$$

where $\text{conv}_{3 \times 3}(\cdot)$ denotes the $3 \times 3$ convolution and $\text{Loss}(\cdot)$ denotes the cross entropy loss function.

Now we explain how to get the subsequent segmentation map $P_m^{RU_k}$ $(k > 1)$ with larger spatial dimensions. One simple solution is to upsample the previous segmentation map $P_m^{RU_{k-1}}$. However, this upsampled segmentation map will be very coarse. In order to derive a more precise segmentation, we use the idea of skip-connections [19], [4]. The basic idea is to make use of the outputs from one of the convolutional feature layers in the encoder network. Since the convolutional feature layer contains more precise spatial information, we can combine it with the upsampled segmentation map to obtain a refined segmentation map (see Fig. 5 for detailed illustration of our refinement module). In our model, we simply concatenate the outputs from the skip layer with the upsampled decoder layer to create a larger feature map, then use $3 \times 3$ convolution across the channels to convert the channel dimension to $C$. For example, the label map $P_m^{RU_1}$ is obtained from upsampled $P_m^G$ and the convolutional feature map $f_5(I)$ (see Fig. 4 for details on these skip connections). We can then define a loss function on this (larger) segmentation map by comparing $P_m^{RU_k}$ with the resized ground-truth segmentation map $R_k(Y)$ of the corresponding size. These operations can be summarized as follows:

$$P_m^{RU_k} = \text{conv}_{3\times3}\Big(\text{concat}\big(\mathbb{U}\big(P_m^{RU_{k-1}}\big), f_{7-k}(I)\big)\Big) \quad (2)$$

$$\ell_{k+1} = \text{Loss}\Big(R_k(Y), \mathbb{S}\big(P_m^{RU_k}\big)\Big), \quad \text{where } k = 1, .., 5 \quad (3)$$

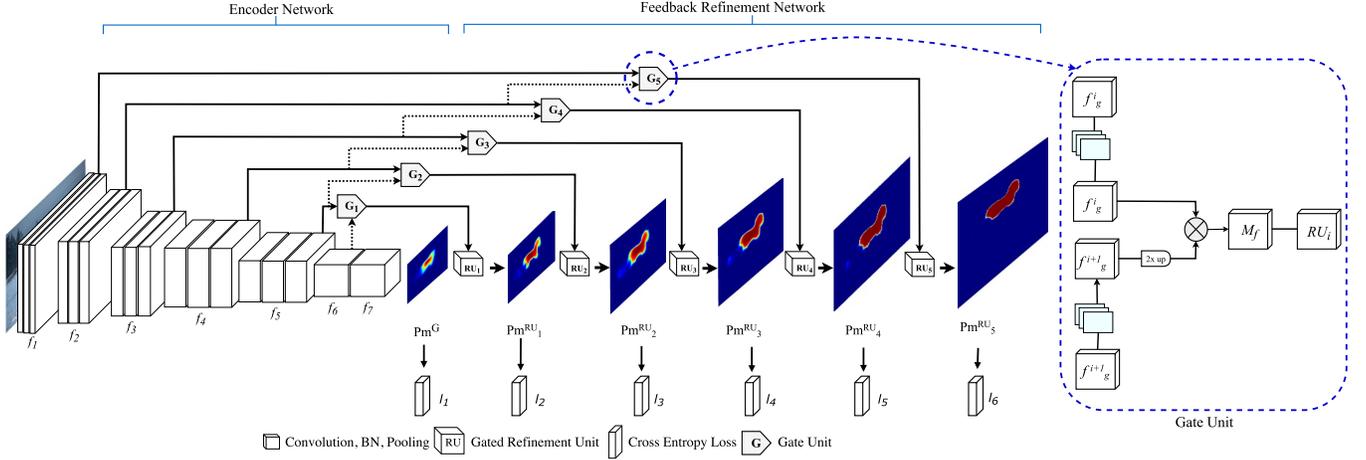

Fig. 4. Overview of our Gated Feedback Refinement Network (G-FRNet). We use feature maps with different spatial dimensions produced by the encoder ($f_1$, $f_2$,..., $f_7$) to reconstruct a small (i.e. coarse) label map $Pm^G$. The decoder progressively refines the label map by adding details from feature maps in the encoder network. At each stage of decoding, a refinement unit ($RU_1$, $RU_2$,..., $RU_5$) produces a new label map with larger spatial dimensions by taking information from the previous label map and encoder layers as inputs (denoted by the edge connecting $G_i$ and $RU_i$). The main novelty of the model is that information from earlier encoder layers passes through a gate unit before being forwarded to the decoder. We use standard 2x bilinear upsampling on each class score map before passing it to the next stage refinement module. We also use down-sampled ground-truth label maps to provide supervision ($l_1$, $l_2$, ..., $l_6$) at each decoding stage.

## V. GATED FEEDBACK REFINEMENT NETWORK

In this section, we describe our proposed network called *Gated Feedback Refinement Network* (G-FRNet) for semantic segmentation.

G-FRNet is built upon LRN. LRN uses skip connections to directly connect two layers in a network, i.e. an encoder layer to a decoder layer. For example, in the network architecture of Fig. 4, a traditional skip connection might connect $f_5$ with $Pm^{RU_1}$. Although this allows the network to pass finer detailed information from the early encoder layers to the decoder, it may degrade the quality of predictions. As mentioned earlier, the categorical ambiguity in early encoder layers may be passed to the decoder.

The main novelty G-FRNet is that we use a gating mechanism to modulate the information being passed via the skip connections. For example, say we want to have a skip connection to pass information from the encoder layer $f_5$ to the decoder layer $Pm^{RU_1}$. Instead of directly passing the feature map $f_5$, we first compute a gated feature map $f_g$ based on $f_5$ and an encoder layer above (i.e. $f_6$ in Fig. 4). The intuition is that $f_6$ contains information that can help resolve ambiguity present in $f_5$. For instance, some of the neurons in $f_6$ might fire on image patches that look like an animal (either cow or horse). This ambiguity about categories (cow vs. horse) cannot be resolved by $f_5$ alone since the receptive field corresponding to this encoder layer might not be large or discriminative enough. But the encoder layer (e.g. $f_6$) above may not be subject to these limitations and provide unambiguous confidence for the correct category. By computing the gated feature map from $f_5$ and $f_6$, categorical ambiguity can be filtered out before reaching the decoding stage where spatial precision is recovered. Fig. 2 provides an example of categorical ambiguity.

The gated feature map from $G_1$ contains information about finer image details. We then combine it with the coarse label map $Pm^G$ to produce an enlarged label map $Pm^{RU_1}$. We repeat this process to produce progressively larger label maps ($Pm^{RU_1}$, $Pm^{RU_2}$, $Pm^{RU_3}$, $Pm^{RU_4}$, $Pm^{RU_5}$).

In the following sections we discuss Gate Unit (Sec. V-A), Gated Refinement Unit (Sec. V-B) and Stage-wise supervision in detail (Sec. V-C).

### A. Gate Unit

Previous work [34] proposed refinement across different levels by combining convolution features from earlier layers. Instead of combining convolution features with coarse label maps directly, we introduce gate units to control the information passed on. The gate units are designed to control the information passed on by modulating the response of encoder layers for each spatial region in a top-down manner. Fig. 4 (right) illustrates the architecture of a gate unit.

The gate unit takes two consecutive feature maps $f_g^i$ and $f_g^{i+1}$ as its input. The features in $f_g^i$ are high-resolution with smaller receptive fields (i.e. small context), whereas features in $f_g^{i+1}$ are of low-resolution with larger receptive fields (i.e. large context). A gate unit combines $f_g^i$ and $f_g^{i+1}$ to generate rich contextual information. Alternative approaches apply a refinement process straight away that combines convolution features (using skip connections [4]) with coarse label maps through concatenation to generate a new label map. In this case, it is less likely that the model will take full advantage of the contribution of higher resolution feature maps if they carry activation that is ambiguous with respect to class. As a result, skip connections alone have inherent limitations in discerning missing spatial details. Therefore, unlike skip connections we first obtain a gated feature map before passing on the higher resolution encoding to the refinement module. As a result, contextual features will be assigned higher gate values and retain their activation for each successive stage of refinement, while irrelevant/noise regions will be suppressed.

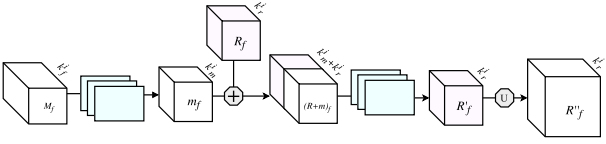

Fig. 5. Detailed overview of a Gated Refinement Unit. The refinement unit is unfolded here for $i^{th}$ stage. The refinement module (similar to [40]) is composed of convolution, batch normalization, concatenation, and upsampling.

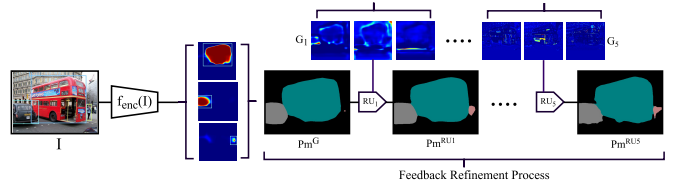

Fig. 6. Visualization of hierarchical gated refinement scheme. The refinement process integrates higher-frequency details with the lower resolution label map at each stage. Class-wise activation maps for each gate are shown as heatmaps.

We now explain how we obtain a gated feature map from a gate unit. Given that the dimensions of feature map $f_g^i$ and the gate control input $f_g^{i+1}$ might not be the same, we first use a transformation function $T_f : \Re^\varepsilon \mapsto \Re^\chi$ to map $f_g^{i+1}$ to $f_{g'}^{i+1}$. In the tranformation process, a sequence of operations is carried out on $f_g^i$ and $f_g^{i+1}$ followed by a element-wise product. Firstly, we apply a $3 \times 3$ convolution with batch normalization and ReLU to both feature maps. After these operations, let $c_g^i$ and $c_g^{i+1}$ be the number of channels in $f_g^i$ and $f_g^{i+1}$ such that $c_g^i = c_{g'}^{i+1}$. $f_g^{i+1}$ is then upsampled by a factor of 2 to produce a new feature map $f_{g'}^{i+1}$ whose spatial dimensions match $f_g^i$. We obtain the $i^{th}$ stage gated (from gate $G_i$ in Fig. 4) feature map $M_f$ from the element-wise product between $f_g^i$ and $f_{g'}^{i+1}$. Finally, the resultant feature map $M_f$ is fed to the gated refinement unit (see Sec. V-B). The formulation of obtaining a gated feature map $M_f$ from gate unit $G_i$ can be written as follows:

$$v_i = T_f(f_g^{i+1}), u_i = T_f(f_g^i), M_f = v_i \odot u_i \quad (4)$$

where $\odot$ denotes element-wise product. As gate units are integrated with the end-to-end network, the parameters could be trained with the back-propagation (BP) algorithm.

### B. Gated Refinement Unit

Fig. 5 shows in detail the architecture of our gated refinement unit (see $RU$ in Fig. 4). Each refinement unit $RU^i$ takes a coarse label map $R_f$ with channel $k_r^i$ (generated at $(i-1)^{th}$ stage of the FRN) and gated feature map $M_f$ as its input. $RU$s learn to aggregate information and generate a new label map $R_f'$ with larger spatial dimensions through the following sequence of operations: First, we apply a $3 \times 3$ convolution followed by a batch normalization layer on $M_f$ to obtain a feature map $m_f$ with channel $k_m^i$. In our model configuration, $k_m^i = k_r^i = C$ where $C$ is the number of possible labels. Next, $m_f$ is concatenated with the prior stage label map $R_f$, producing feature map $(R+m)_f$ with $k_m^i + k_r^i$ channels. There are two reasons behind making $k_m^i = k_r^i$.

First, the channel dimension of the feature map obtained from the encoder is typically very large (i.e. $c_g^i \gg k_r^i$). So directly concatenating $R_f$ with a feature map containing a larger number of channels is computationally expensive. Second, concatenating two feature maps having a large difference in the number of channels risks dropping signals from the representation with fewer channels. Finally, the refined label map $R_f'$ is generated by applying a $3 \times 3$ convolution. Note that $R_f'$ is the $i^{th}$ stage prediction map. The prediction map $R_f'$ is upsampled by a factor of 2 and fed to the next stage $(i+1)^{th}$ gated refinement unit. These operations can be summarized as follows:

$$m_f = \mathbb{B}(\mathbb{C}_{3\times3}(M_f)), \gamma = m_f \oplus R_f, R_f' = \mathbb{C}_{3\times3}(\gamma) \quad (5)$$

where $\mathbb{B}(.)$, $\mathbb{C}(.)$, and $\oplus$ refer to batch normalization, convolution, and concatenation respectively.

### C. Stage-wise Supervision

Our network produces a sequence of label maps with increasing spatial dimensions at the decoder stage, although we are principally interested in the label map at the last stage of the decoding. Label maps produced at earlier stages of decoding might provide useful information as well and allow for supervision earlier in the network. Following [40], we adopt the idea of deep supervision [30] in our network to provide stage-wise supervision on predicted dense label maps. In more specific terms, let $I \in \mathbb{R}^{h \times w \times d}$ be a training sample with ground-truth mask $\eta \in \mathbb{R}^{h \times w}$. We obtain $k$ resized ground-truth maps $(R_1, R_2, ...., R_k)$ by resizing $\eta$. We define a loss function $l_i$ (pixel-wise cross entropy loss is used) to measure the difference between resized ground-truth $R_i(\eta)$ and the predicted label map at each stage of decoding. We can write these operations as follows:

$$l_k = \begin{cases} \xi\Big(R_i(\eta), Pm^G\Big) & \text{if } i = 1 \\ \xi\Big(R_i(\eta), Pm^{RU_i}\Big) & \text{otherwise} \end{cases} \quad (6)$$

where $\xi$ denotes weighted cross-entropy loss which is defined by the following:

$$\xi = -\lambda \sum_{t=1}^{N} \sum_{t=1}^{C} t_i \log(x_i), \qquad \ell = \sum_{k=1}^{6} l_k \quad (7)$$

$\lambda$ is the class balancing frequency or stage-specific weight factor on a per-pixel term basis. The loss function $\ell$ in our network is the summation of cross-entropy losses at various stages of refinement network. The network is trained using back-propagation to optimize this loss. Fig. 6 illustrates the effectiveness of the gated refinement scheme. We can see that the refinement scheme progressively improves the spatial details of dense label maps. It also shows that the top convolution layer (conv7 in our encoder network) can predict a coarse label map without capturing finer image details. The feedback refinement network is able to recover missing details (e.g. the boundaries of the bus and the car) in the coarse label map.

## VI. Experiments

[ht] In this section, we first provide implementation details (Sec. VI-A). Then we present experimental results on five challenging dense labeling benchmark datasets: Cambridge Driving Labeled Video (CamVid) (Sec. VI-B), PASCAL VOC 2012 (Sec. VI-C), Horse-Cow Parsing (Sec. VI-D), PASCAL-Person-Part (Sec. VI-E), and SUN-RGBD (Sec. VI-F).

### A. Implementation Details

We have implemented our network using Caffe [45] on a single Titan X GPU. Pre-trained VGG-16 [1] parameters are used to initialize the convolution layers in the encoder network (i.e. $conv1$ to $conv5$ layer). Other convolution layers' parameters are randomly assigned based on Xavier initialization. Randomly cropped patches of size $(h_{min} \times w_{min})$ are fed into the network. We set $(h_{min} \times w_{min})$ to $320 \times 320$ for Pascal VOC and $360 \times 480$ for CamVid and Horse-Cow parsing datasets. For the PASCAL VOC 2012 dataset, we normalize the data using VGG-16 mean and standard deviation. We employ pixel-wise cross entropy loss (with equal weights) as the objective function to be optimized for all the semantic categories. For the CamVid dataset, since data is not balanced we use weighted cross entropy loss following previous work [6]. The weights are computed using the class balancing technique proposed in [26].

Since the gated refinement modules can handle inputs of any size, we are able to test our network with original image size. The spatial resolution of the final segmentation map is therefore the same as the test image.

To demonstrate the merit of individual component of our model, we also perform an ablation study by comparing with the following variants of our proposed model:

**LRN:** Label Refinement Network that only uses upper layer features for refinement.

**G-FRNet:** Gated Feedback refinement network with gating mechanism. The major difference between G-FRNet and LRN is the gate units. To demonstrate the real impact of gate units we consider LRN as our base model and report experimental results for both models in all of our comparisons.

**G-FRNet-Res101:** Gated Feedback refinement network where the encoder/base model (VGG-16) is replaced with dilated ResNet-101 [22]. Inspired by [22], we use the "poly" learning rate policy defined by $(1 - \frac{iter}{maxiter})^{power}$ when training G-FRNet-Res101. The learning rate of the newly added layers are initialized as $2.5 \times 10^{-3}$ and that of other previously learned layers initialized as $2.5 \times 10^{-4}$. We train the ResNet-101 based models using stochastic gradient descent with a batch size of 1, momentum of 0.9, and weight decay of 0.0005.

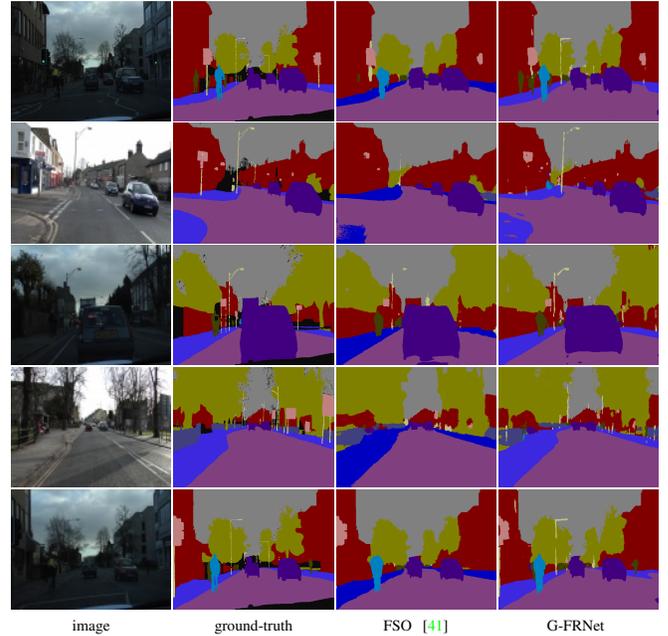

Fig. 7. Qualitative results on the CamVid dataset. G-FRNet is capable of retaining the shape of smaller and finer object categories (e.g. column-pole, side-walk, bicyclist, and sign-symbols) accurately compared to FSO [41].

### B. CamVid

The Cambridge-driving Labeled Video (CamVid) dataset [44] consists of 701 high resolution video frames extracted from a video footage recorded in a challenging urban setting. Ground-truth labels are annotated according to one of 32 semantic categories. Following [41], [6], [46], we consider 11 larger semantic classes (road, building, sky, tree, sidewalk, car, column-pole, fence, pedestrian, bicyclist, and sign-symbol) for evaluation. We split the dataset into training, validation, and test sets following [47]. Finally, we have 367 training images, 100 validation images, and 233 test images. In order to make our experimental settings comparable to previous works [41], [24], [46], [6], we downsample the images in the dataset by a factor of 2 (i.e. $480 \times 360$). Table I shows the results of our model and comparisons with other state-of-the-art approaches on this dataset, demonstrating that we achieve state-of-the-art results on this dataset. G-FRNet has produced significantly better performance over other recently developed segmentation network architectures including SegNet [6], DilatedNet [24], FSO [41], DeepLab [17], DenseNet [43], etc. For each method, we report the category-wise IoU score and mean IoU score. LRN [40] outperforms SegNet [6] by more than 11% (in terms of mean IoU) while our approach (i.e. G-FRNet) achieves an accuracy gain of 6% when compared with DeepLab [17] and by almost 2% over Dilation [24] and FSO [41]. G-FRNet[††] further improves the performance and yields 68.8% mean IoU.

Fig. 7 shows some qualitative results on this dataset. We can see that our model is especially accurate for challenging object categories, such as column-pole, sidewalk, bicyclist, and sign-symbols compared to [41].

| Method | Building | Tree | Sky | Car | Sign | Road | Pedestrian | Fence | Pole | Sidewalk | Bicyclist | Mean IoU |
|---|---|---|---|---|---|---|---|---|---|---|---|---|
| | w/o ConvNet | | | | | | | | | | | |
| SuperParsing [41] | 70.4 | 54.8 | 83.5 | 43.3 | 25.4 | 83.4 | 11.6 | 18.3 | 5.2 | 57.4 | 8.9 | 42.0 |
| TextonBoost + FSO [41] | 74.4 | 71.8 | 91.6 | 64.9 | 27.7 | 91.0 | 33.8 | 34.1 | 16.8 | 73.9 | 27.6 | 55.2 |
| | with ConvNet | | | | | | | | | | | |
| Bayesian SegNet [42] | n/a | | | | | | | | | | | 63.1 |
| DeconvNet [7] | n/a | | | | | | | | | | | 48.9 |
| SegNet [6] | 68.7 | 52 | 87 | 58.5 | 13.4 | 86.2 | 25.3 | 17.9 | 16.0 | 60.5 | 24.8 | 50.2 |
| FCN-8s [4] | 77.8 | 71.0 | 88.7 | 76.1 | 32.7 | 91.2 | 41.7 | 24.4 | 19.9 | 72.7 | 31.0 | 57.0 |
| DeepLab-LargeFOV [17] | 81.5 | 74.6 | 89.0 | 82.2 | 42.3 | 92.2 | 48.4 | 27.2 | 14.3 | 75.4 | 50.1 | 61.6 |
| Dilation [24] | 82.6 | 76.2 | 89.9 | 84.0 | 46.9 | 92.2 | 56.3 | 35.8 | 23.4 | 75.3 | 55.5 | 65.3 |
| Dilation + FSO + DiscreteFlow [41] | 84.0 | 77.2 | 91.3 | 85.6 | 49.9 | 92.5 | 59.1 | 37.6 | 16.9 | 76.0 | 57.2 | 66.1 |
| DenseNet103 [43] | 83.0 | 77.3 | 93.0 | 77.3 | 43.9 | 94.5 | 59.6 | 37.1 | 37.8 | 82.2 | 50.5 | 66.9 |
| LRN [40] | 78.6 | 73.6 | 76.4 | 75.2 | 40.1 | 91.7 | 43.5 | 41.0 | 30.4 | 80.1 | 46.5 | 61.7 |
| **G-FRNet** | 82.5 | 76.8 | 92.1 | 81.8 | 43.0 | 94.5 | 54.6 | 47.1 | 33.4 | 82.3 | 59.4 | 68.0 |
| **G-FRNet††** | 83.7 | 77.8 | 92.3 | 83.6 | 44.7 | 94.6 | 58.4 | 45.2 | 34.7 | 83.2 | 58.1 | 68.8 |

TABLE I

QUANTITATIVE RESULTS ON THE CAMVID DATASET [44]. WE REPORT PER-CLASS IOU AND MEAN IOU FOR EACH METHOD. WE SPLIT METHODS INTO TWO CATEGORIES DEPENDING ON WHETHER OR NOT THEY USE CONVNET. NOT SURPRISINGLY, CONVNET-BASED METHODS TYPICALLY OUTPERFORM NON-CONVNET METHODS. OUR APPROACH ACHIEVES STATE-OF-THE-ART RESULTS ON THIS DATASET. NOTE THAT THE IMPROVEMENTS ON SMALLER AND FINER OBJECTS ARE PARTICULARLY PRONOUNCED FOR OUR MODEL. THE PRECEDING SYMBOL †† INDICATES THAT THE NETWORK USED A PRE-TRAINED MODEL WHICH IS TRAINED ON SYNTHETIC DATA.

### C. PASCAL VOC 2012

PASCAL VOC 2012 [48] is a challenging dataset for semantic segmentation. This dataset consists of 1,464 training images and 1,449 validation images of 20 object classes (plus the background class). There are 1,456 test images for which ground-truth labels are not publicly available. We obtained results on the test set by submitting our final predictions to the evaluation server. Following prior work [17], [14], [6], we augment the training set with extra labeled PASCAL VOC images from [49]. In the end, we have 10,582 labeled training images.

In Table II, we compare our results on the validation set with previous works. G-FRNet + CRF achieves best result with 71.0% mean IoU accuracy compared to other models based on an encoder-decoder based architecture ([7], [50], [4]). When we switch to a base model that exhibits stronger base performance (e.g. ResNet-101 [22] instead of VGG) our model G-FRNet-Res101 + CRF achieves 77.8% mean IoU which is very competitive compared to recent ResNet based state-of-the-art methods. In order to examine the importance of deep supervision (Sec. V-C), we conduct an additional experiment (G-FRNet-Res101 (w/o DS)) where we remove the stage-wise losses ($l_1, l_2, l_3, l_4, l_5$) (see Fig. 4) and just optimize the network using the final loss $l_6$. We obtain the overall performance of 71.8% mIoU which is 4.7% lesser than the model with deep supervision (G-FRNet-Res101). This strongly demonstrate the importance of stage-wise supervision strategy in the refinement based architectures.

In order to evaluate on test set, we train our model on both the training and validation set. Table III shows quantitative results of our method on the test set. We achieve very competitive performance compared to other baselines. LRN [40]

| Method | Mean IoU (%) |
|---|---|
| DeepLab-CRF-LargeFOV [17] | 67.6 |
| DeepLab-MSc-CRF-LargeFOV [17] | 68.7 |
| FCN [4] | 61.3 |
| DeconvNet (w/o object proposals) [7] | — |
| DeconvNet (w/ object proposals) [7] | 67.1 |
| DeconvNet [7] | 67.1 |
| Attention [51] | 71.4 |
| DeepLabv2 [22] | 77.7 |
| LRN [40] | 62.8 |
| G-FRNet | 68.7 |
| G-FRNet + CRF | 71.0 |
| G-FRNet-Res101 (w/o DS) | 71.8 |
| G-FRNet-Res101 | 76.5 |
| **G-FRNet-Res101 + CRF** | 77.8 |

TABLE II

COMPARISON OF DIFFERENT METHODS ON PASCAL VOC 2012 VALIDATION SET. NOTE THAT DECONVNET [7] RESULT IS TAKEN FROM [50].

achieves 64.2% mean IoU which outperforms FCN [4] and SegNet [6]. Our proposed approach G-FRNet improves the mean IoU accuracy by 4%. Many existing works (e.g. [17], [7], [22], [51]) use a CRF model [53] as a post-processing to improve the performance. When we apply CRF on top of our final prediction (G-FRNet + CRF), we further improve the mean IoU to 70.4% on the test set. G-FRNet-Res101 (with CRF) further improves the performance and yields 79.3% mean IoU on test set which is very competitive compared to existing state-of-the-art approaches. A link to the results from the benchmark site is provided [1].

In recent years, many semantic segmentation methods have

---

[1] http://host.robots.ox.ac.uk:8080/anonymous/HU5Y96.html

| method | aero | bike | bird | boat | bottle | bus | car | cat | chair | cow | table | dog | horse | mbike | person | plant | sheep | sofa | train | tv | **mIoU** |
|---|---|---|---|---|---|---|---|---|---|---|---|---|---|---|---|---|---|---|---|---|---|
| Hypercolumn [19] | 68.4 | 27.2 | 68.2 | 47.6 | 61.7 | 76.9 | 72.1 | 71.1 | 24.3 | 59.3 | 44.8 | 62.7 | 59.4 | 73.5 | 70.6 | 52.0 | 63.0 | 38.1 | 60.0 | 54.1 | 59.2 |
| FCN-8s [4] | 76.8 | 34.2 | 68.9 | 49.4 | 60.3 | 75.3 | 74.7 | 77.6 | 21.4 | 62.5 | 46.8 | 71.8 | 63.9 | 76.5 | 73.9 | 45.2 | 72.4 | 37.4 | 70.9 | 55.1 | 62.2 |
| SegNet [6] | 74.5 | 30.6 | 61.4 | 50.8 | 49.8 | 76.2 | 64.3 | 69.7 | 23.8 | 60.8 | 54.7 | 62.0 | 66.4 | 70.2 | 74.1 | 37.5 | 63.7 | 40.6 | 67.8 | 53.0 | 59.1 |
| Zoom-Out [20] | 85.6 | 37.3 | 83.2 | 62.5 | 66.0 | 85.1 | 80.7 | 84.9 | 27.2 | 73.2 | 57.5 | 78.1 | 79.2 | 81.1 | 77.1 | 53.6 | 74.0 | 49.2 | 71.7 | 63.3 | 64.4 |
| DeconvNet[7] | 87.8 | 41.9 | 80.6 | 63.9 | 67.3 | 88.1 | 78.4 | 81.3 | 25.9 | 73.7 | 61.2 | 72.0 | 77.0 | 79.9 | 78.7 | 59.5 | 78.3 | 55.0 | 75.2 | 61.5 | 70.5 |
| DeepLab [17] | 84.4 | 54.5 | 81.5 | 63.6 | 65.9 | 85.1 | 79.1 | 83.4 | 30.7 | 74.1 | 59.8 | 79.0 | 76.1 | 83.2 | 80.8 | 59.7 | 82.2 | 50.4 | 73.1 | 63.7 | 71.6 |
| Dilation [24] | 91.7 | 39.6 | 87.8 | 63.1 | 71.8 | 89.7 | 82.9 | 89.8 | 37.2 | 84.0 | 63.0 | 83.3 | 89.0 | 83.8 | 85.1 | 56.8 | 87.6 | 56.0 | 80.2 | 64.7 | 75.3 |
| Attention [51] | 93.2 | 41.7 | 88.0 | 61.7 | 74.9 | 92.9 | 84.5 | 90.4 | 33.0 | 82.8 | 63.2 | 84.5 | 85.0 | 87.2 | 85.7 | 60.5 | 87.7 | 57.8 | 84.3 | 68.2 | 76.3 |
| DeepLabv2 [22] | 92.6 | 60.4 | 91.6 | 63.4 | 76.3 | 95.0 | 88.4 | 92.6 | 32.7 | 88.5 | 67.6 | 89.6 | 92.1 | 87.0 | 87.4 | 63.3 | 88.3 | 60.0 | 86.8 | 74.5 | 79.7 |
| PSPNet [52] | 95.8 | 72.7 | 95.0 | 78.9 | 84.4 | 94.7 | 92.0 | 95.7 | 43.1 | 91.0 | 80.3 | 91.3 | 96.3 | 92.3 | 90.1 | 71.5 | 94.4 | 66.9 | 88.8 | 82.0 | **85.4** |
| LRN [40] | 79.3 | 37.5 | 79.7 | 47.7 | 58.3 | 76.5 | 76.1 | 78.5 | 21.9 | 67.7 | 47.6 | 71.2 | 69.1 | 82.1 | 77.5 | 46.8 | 70.1 | 40.3 | 71.5 | 57.4 | 64.2 |
| G-FRNet | 84.8 | 39.6 | 80.3 | 53.9 | 58.1 | 81.7 | 78.2 | 78.9 | 28.8 | 75.3 | 55.2 | 74.7 | 75.5 | 81.9 | 79.7 | 51.7 | 76.3 | 43.2 | 80.1 | 62.3 | 68.2 |
| G-FRNet + CRF | 87.7 | 42.9 | 85.4 | 51.6 | 61.0 | 82.9 | 81.7 | 81.6 | 29.1 | 79.3 | 56.1 | 77.6 | 78.6 | 84.6 | 81.6 | 52.8 | 79.0 | 45.0 | 82.1 | 64.1 | 70.4 |
| **G-FRNet-Res101** | 91.4 | 44.6 | 91.4 | 69.2 | 78.2 | 95.4 | 88.9 | 93.3 | 37.0 | 89.7 | 61.4 | 90.0 | 91.4 | 87.9 | 87.2 | 63.8 | 89.4 | 59.9 | 87.0 | 74.1 | 79.4 |

TABLE III

QUANTITATIVE RESULTS IN TERMS OF MEAN IoU ON PASCAL VOC 2012 TEST SET. NOTE THAT G-FRNET-RES101 INCLUDES CRF.

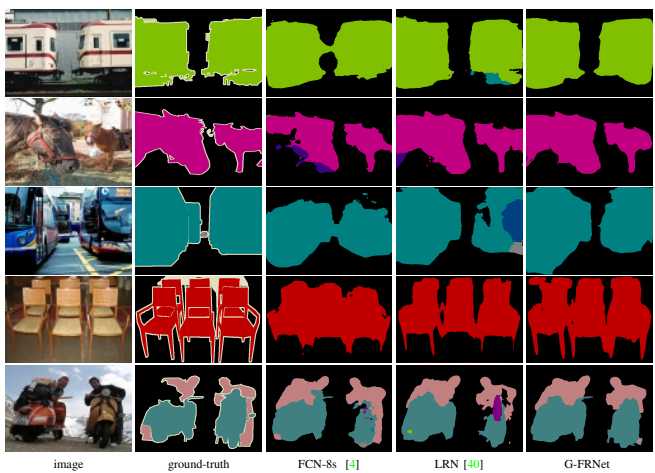

Fig. 8. Qualitative results on PASCAL VOC 2012 validation set. G-FRNet again produces better labeling when compared to [4] and LRN [40].

been proposed based on PASCAL VOC 2012 which are increasingly more precise in terms of IoU measure, and also introduce significant additional model complexity. However, there are only few recent methods [7], [6] that use a simpler encoder-decoder architecture for this problem, and it is most natural to compare our approach directly with this related family of models. Unlike other baseline methods, we obtain these results without employing any performance enhancing techniques, such as using object proposals [7] and multi-stage training [7]. It is worth noting that while the proposed model is shown to be highly capable across several datasets, a deeper ambition of this paper is to demonstrate the power of basic information routing mechanisms provided by gating in improving performance. The encoder-decoder based architecture provides a natural vehicle for this demonstration. It is expected that a wide variety of networks that abstract away spatial precision in favor of a more complex pool of features may benefit from installing similar logic to the proposed gating mechanism.

### D. Horse-Cow Parsing Dataset

To further confirm the value and generality of our model for dense labeling problems, we also evaluate our model on object parts parsing dataset introduced in [54]. This dataset contains images of horses and cows only, which are manually selected from the PASCAL VOC 2010 benchmark [48] based on most observable instances. The task is to label each pixel according to whether this pixel belongs to one of the body parts (head, leg, tail, body). We split the dataset following [54] and obtain 294 training images and 227 test images. We compare the performance of our model with state-of-the art methods including the most recent method LG-LSTM [55].

Table IV shows the performance of our models and comparisons with other baseline methods. LRN achieves competitive performance (65.89% and 62.53% mean IoU on horse and cow parsing respectively) whereas the proposed G-FRNet architecture makes further improvement and outperforms all the baselines in terms of mean IoU. The results reach 70.83% mean IoU for horse parsing and 65.35% for cow parsing. We also provide qualitative results in Fig. 9. The superior performance achieved by our model indicates that integrating gate units in the refinement process is very effective in capturing complex contextual patterns within images which play a critical role in distinguishing and segmenting different localized semantic parts of an instance.

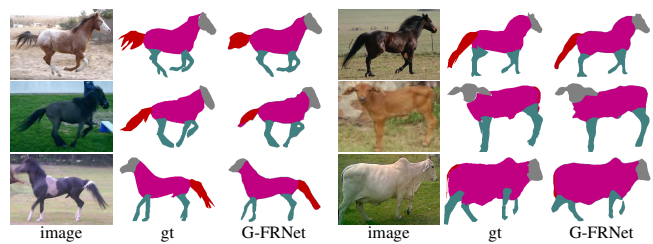

Fig. 9. Qualitative results on Horse-Cow parsing dataset [54].

| Method | Horse | | | | | | Cow | | | | | |
|---|---|---|---|---|---|---|---|---|---|---|---|---|
| | Bkg | head | body | leg | tail | IoU | Bkg | head | body | leg | tail | IoU |
| SPS [54] | 79.14 | 47.64 | 69.74 | 38.85 | - | - | 78.0 | 40.55 | 61.65 | 36.32 | - | - |
| SPS- Guidance [56] | 76.0 | 55.0 | 52.4 | 46.8 | 37.2 | 50.3 | 69.7 | 57.6 | 62.7 | 38.5 | 11.8 | 48.03 |
| HC [19] | 85.71 | 57.30 | 77.88 | 51.93 | 37.10 | 61.98 | 81.86 | 55.18 | 72.75 | 42.03 | 11.04 | 52.57 |
| JPO [57] | 87.34 | 60.02 | 77.52 | 58.35 | 51.88 | 67.02 | 85.68 | 58.04 | 76.04 | 51.12 | 15.00 | 57.18 |
| DeepLab-LargeFoV [17] | 87.44 | 64.45 | 80.70 | 54.61 | 44.03 | 66.25 | 86.56 | 62.76 | 78.42 | 48.83 | 19.97 | 59.31 |
| LG - LSTM [55] | 89.64 | 66.89 | 84.20 | 60.88 | 42.06 | 68.73 | 89.71 | 68.43 | 82.47 | 53.93 | 19.41 | 62.79 |
| LRN [40] | 90.11 | 53.23 | 81.57 | 56.50 | 48.03 | 65.89 | 90.30 | 64.41 | 81.52 | 53.44 | 23.03 | 62.53 |
| **G-FRNet** | 91.79 | 60.44 | 84.37 | 64.07 | 53.47 | 70.83 | 91.48 | 69.26 | 84.10 | 57.58 | 24.31 | 65.35 |

TABLE IV
COMPARISON OF OBJECT PARSING PERFORMANCE WITH STATE-OF-THE-ART METHODS ON HORSE-COW PARSING DATASET [54].

### E. PASCAL-Person-Part

We further carry out experiments on the PASCAL-Person-Part dataset, a subset of PASCAL VOC 2010 images introduced in [54]. This dataset consists of humans images with variety of poses and scales. It includes pixel-level annotations for six person parts: Head, Torso, Upper/Lower Arms, Upper/Lower Legs and the Background. Following [22], [35], we use only those PASCAL VOC images that contains at least one person. The dataset has 1717 training images and 1818 test images. We evaluate only our best network G-FRNet† on this dataset.

We report segmentation results of PASCAL-Person-Part dataset in Table V. We also compare our results with other state-of-the-art methods. The results clearly demonstrate the effectiveness of our network. Qualitative examples of our object parsing results on this dataset are shown in Fig. 10.

| Method | Mean IoU (%) |
|---|---|
| DeepLab [17] | 51.8 |
| LG-LSTM [55] | 57.97 |
| Attention [51] | 56.39 |
| HAZN [58] | 57.54 |
| Graph LSTM [59] | 60.16 |
| DeepLabv2 (ResNet-101) [22] | **64.94** |
| LRN-Res101 [40] | 60.75 |
| **G-FRNet-Res101** | 64.61 |

TABLE V
COMPARISON OF OBJECT PARSING RESULTS WITH OTHER STATE-OF-THE-ART RESULTS ON PASCAL PERSON-PART DATASET [54].

### F. SUN RGB-D

We also evaluate our model on the SUN RGB-D dataset [60] which contains 5,285 training and 5,050 test images. The images consist of indoor scenes of varying resolution and aspect ratio. There are 37 indoor scene classes with corresponding segmentation labels (background is not considered as a class and is ignored during training and testing). The segmentation labels are instance-wise, i.e. multiple instances of same class in an image have different labels. We convert the ground-truth labels into class-wise segmentation labels so that all instances of the same class have the same corresponding label. Although the dataset also contains depth information, we only use the RGB images to train and test our model. Quantitative results on this dataset are shown in Table VI. Our LRN model achieves better mean IoU than other baselines on this dataset. G-FRNet-Res101 further improves the performance and yields 36.86% mean IoU. We show some qualitative results in Fig. 11.

| Method | Pixel Acc. | Mean Acc. | Mean IoU |
|---|---|---|---|
| FCN-8s [4] | 68.18 | 38.41 | 27.39 |
| DeepLab [17] | 71.90 | 42.21 | 32.08 |
| DeconvNet [7] | 66.13 | 33.28 | 22.57 |
| SegNet [6] | 70.3 | 35.6 | 26.3 |
| Bayesian SegNet [42] | 72.63 | 44.76 | 30.7 |
| SegNet + DS | 71.3 | 49.2 | 31.2 |
| LRN [40] | 72.5 | 46.8 | 33.1 |
| **G-FRNet-Res101** | 75.33 | 47.49 | **36.86** |

TABLE VI
COMPARISON OF SCENE PARSING RESULTS WITH OTHER STATE-OF-THE-ART RESULTS ON SUN RGB-D DATASET [60].

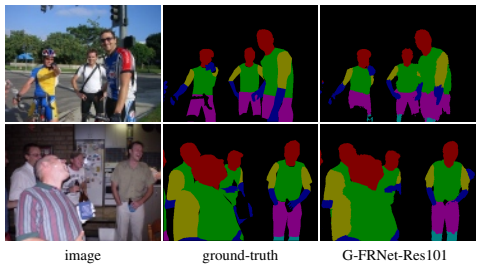

Fig. 10. Qualitative results on the Pascal Person-Part dataset. G-FRNet-Res101 produces output closer to ground-truth.

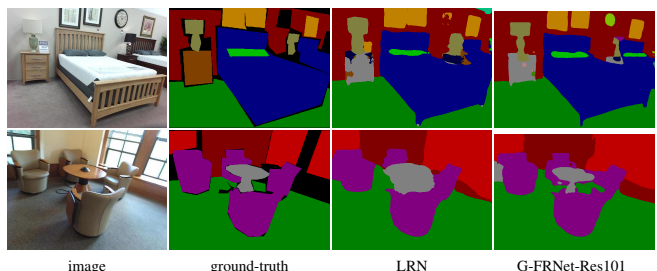

Fig. 11. Qualitative results on the SUN RGB-D dataset.

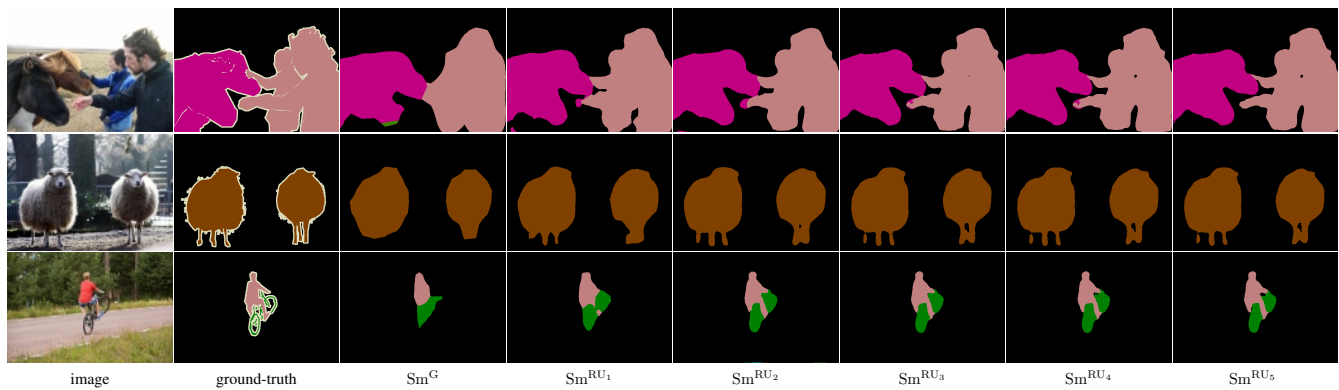

Fig. 12. Stage-wise visualization of semantic segmentation results on PASCAL VOC 2012. For each row, we show the input image, ground-truth, and the prediction map produced at each stage of our feedback refinement network.

### G. Ablation Analysis

To further illustrate the impact of the gated coarse-to-fine refinement, we show segmentation maps produced at different stages (see Fig. 12) in the network. We can see that gated coarse-to-fine refinement scheme progressively improves the details of predicted segmentation maps by recovering the missing parts (e.g., the leg of the person and sheep in the top and $3^{rd}$ row respectively). We perform ablation analysis to demonstrate the benefit of our coarse-to-fine approach and gate units. We first perform a controlled study to isolate the effect of gate units on labeling accuracy. Then we include the gate units and train the model on three different datasets. Table VII and Table VIII show the results of this stage-wise analysis on the datasets used. We can see that, from $\mathrm{Pm^G}$ to $\mathrm{Pm^{RU_5}}$, mean IoU is progressively enhanced in all the datasets. Note that $\mathrm{Pm^G}, .., \mathrm{Pm^{RU_5}}$ are not results of separate stage-wise training. They are obtained from different stages of the feedback refinement network.

The result of $\mathrm{Pm^G}$ is implicitly affected by the supervisions provided at $\mathrm{Pm^{RU_2}}, .., \mathrm{Pm^{RU_5}}$. Note that PASCAL VOC 2012 stage-wise results are without using CRF or ResNet-101. Fig. 13 shows the stage-wise performance (in terms of mean IoU (%)) of G-FRNet and LRN [40] on the datasets used in this paper. Recall that the difference between G-FRNet and LRN are the gate units. From this analysis, it is clear that the inclusion of gate units not only improves the overall performance of the network but also achieves performance gains at each stage of the feedback refinement network. Fig. 14 demonstrates the value of gate units in resolving categorical ambiguity that may arise in the refinement process.

| Stages | CamVid | | PASCAL VOC 2012 | | SUN-RGBD | |
|---|---|---|---|---|---|---|
| | LRN | G-FRNet | LRN | G-FRNet | LRN | G-FRNet |
| $\mathrm{P_m^G}$ | 50.9 | 54.5 | 58.4 | 64.1 | 32.67 | 31.0 |
| $\mathrm{P_m^{RU_1}}$ | 55.5 | 61.3 | 61.6 | 66.6 | 33.91 | 31.6 |
| $\mathrm{P_m^{RU_2}}$ | 59.1 | 65.2 | 61.9 | 68.1 | 34.54 | 32.4 |
| $\mathrm{P_m^{RU_3}}$ | 60.9 | 67.1 | 62.6 | 68.3 | 35.6 | 32.7 |
| $\mathrm{P_m^{RU_4}}$ | 61.4 | 67.8 | 62.5 | 68.6 | 36.31 | 32.8 |
| $\mathrm{P_m^{RU_5}}$ | **61.7** | **68.0** | **62.8** | **68.7** | **36.86** | **33.1** |

TABLE VII
Stage-wise mean IoU on PASCAL VOC 2012 validation set, CamVid, and SUN-RGBD dataset.

| Stages | Horse Parsing | | Cow Parsing | | Pascal-Person-Part | |
|---|---|---|---|---|---|---|
| | LRN | G-FRNet | LRN | G-FRNet | LRN | G-FRNet |
| $\mathrm{P_m^G}$ | 60.6 | 66.42 | 56.22 | 60.35 | 60.51 | 58.79 |
| $\mathrm{P_m^{RU_1}}$ | 64.73 | 68.83 | 60.15 | 62.59 | 62.36 | 59.2 |
| $\mathrm{P_m^{RU_2}}$ | 64.78 | 70.05 | 61.87 | 64.61 | 63.6 | 59.4 |
| $\mathrm{P_m^{RU_3}}$ | 65.78 | 70.70 | 62.46 | 65.15 | 63.97 | 60.10 |
| $\mathrm{P_m^{RU_4}}$ | 65.81 | 70.79 | 62.49 | 65.2 | 64.11 | 60.3 |
| $\mathrm{P_m^{RU_5}}$ | **65.89** | **70.83** | **62.53** | **65.35** | **64.61** | **60.75** |

TABLE VIII
Stage-wise mean IoU on Horse-Cow parsing and Pascal-Person-Part dataset. Pascal-Person-Part stage-wise numbers are reported for G-FRNet-Res101 and LRN-Res101

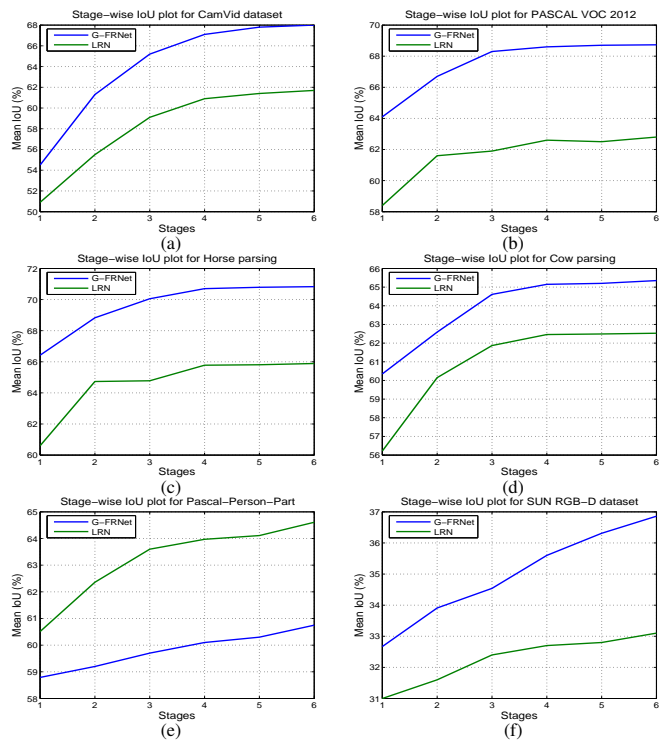

Fig. 13. Comparison of stage-wise mean IoU on (a) CamVid dataset; (b) PASCAL VOC 2012 validation set (c) Horse parsing (d) Cow parsing (e) Pascal-Person-Part and (f) SUN-RGBD dataset between LRN [40] and proposed network G-FRNet. Note that Pascal-Person-Part stage-wise results are using G-FRNet-Res101 and LRN-Res101.

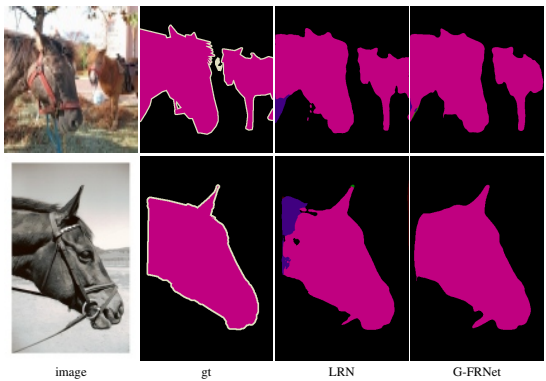

Fig. 14. Illustration of categorical ambiguity that can be resolved by gate units proposed in G-FRNet.

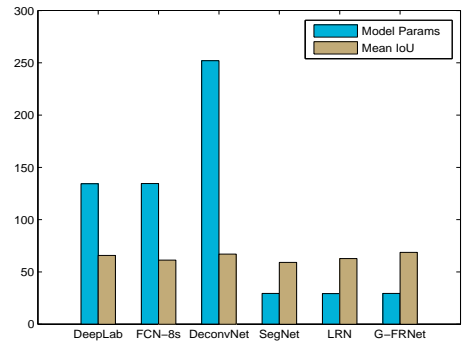

Fig. 16. Analysis on the number of model parameters (in millions) and the mean IoU (%) on PASCAL VOC 2012 validation set for different methods. The rightmost method is our proposed model, which achieves best performance, even with considerably fewer parameters, and a more parsimonious model structure.

### H. Exploration Studies

From the qualitative results shown in Fig. 7, Fig. 9, Fig. 10, and Fig. 11 we can see that our predictions are more precise and semantically meaningful than the baselines. For example, smaller regions (e.g. tail) in the horse-cow parsing dataset and thinner objects (e.g. column-pole, pedestrian, sign-symbol) in the CamVid dataset can be precisely labeled by G-FRNet. G-FRNet is also capable of effectively handling categories that similar in visual appearance (e.g. horse and cow). Regions with similar appearance (e.g. body parts of horse and cow) can be discriminated by the global contextual guidance provided by the gate units. The local boundaries for different semantic regions are preserved using the low-frequency information from earlier layers.

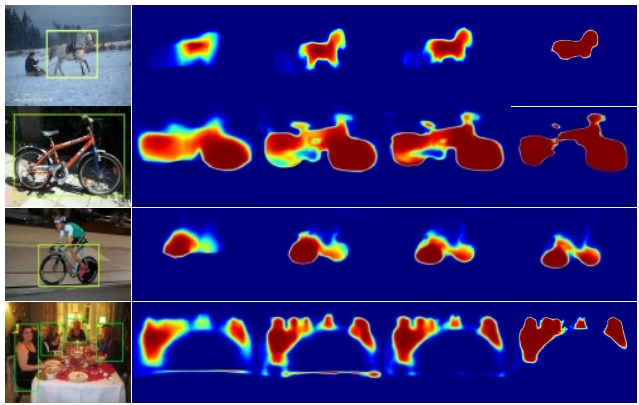

Fig. 15. Class-wise heatmap visualization on PASCAL VOC 2012 validation set images after each stage of refinement. Interestingly, the network gradually aligns itself more precisely with semantic labels, while correcting initially mislabeled regions. The rightmost column shows the heatmap of the final prediction layer.

Fig. 15 shows that prediction quality progressively improves with each successive stage of refinement. In coarse-level predictions, the network is only able to identify some parts of objects or semantic categories. With each stage of gated refinement, missing parts of the object are recovered and mislabeled parts are corrected.

Fig. 16 shows a comparison between different methods in terms of the total number of model parameters and mean IoU (%) on PASCAL VOC 2012 dataset. Although our model has only 12 to 25 percent of the number of parameters of other state-of-the-art methods (FCN [4] and DeconvNet [7]), it achieves very competitive performance. This shows the efficiency of the proposed model despite its simplicity and also the broader value of the proposed gating mechanism.

Fig. 17 shows dense predictions on challenging images of CamVid dataset. The labeling of finer details such as the column pole are improved. This improvement is mainly due to the recurrent connection of the weights in the encoder and decoder networks respectively. In addition, top-down modulation through gate units significantly helps to select relevant low-level features towards recovery of fine spatial details (especially for smaller objects) even after lowering the resolution through several pooling layers.

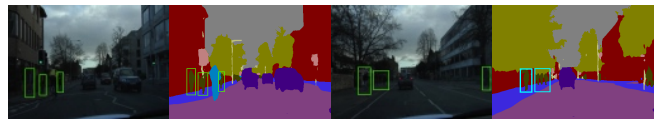

Fig. 17. Qualitative analysis of dense predictions on a few challenging object categories in CamVid dataset. For example, pedestrians are labeled accurately by our model, despite the varying light settings.

Additionally, the value of the gating mechanism is demonstrated in each of the experiments, with its strengths evident in both the qualitative and quantitative results. The LRN method uses the upper layer feature map alone. We reported the result of LRN for all datasets. It is clear that the proposed gating mechanism in G-FRNet significantly improves performance compared with LRN. As higher layers see a larger part of the scene and represent more complex concepts (while also being composed of features from earlier layers) it is natural that subsequent layers are able to resolve ambiguity that preceding layers cannot possibly resolve. Our work is the first that uses a gating mechanism in the encoder-decoder network for dense image labeling. Given the apparent efficacy of the proposed gating mechanism, we expect that this work will inspire significant interest in exploring different gating mechanisms within future work. It is especially noteworthy how powerful this architectural modification is shown to be across a wide array of different datasets, with different properties and labels. With respect to the mechanics of the gating mechanism, we have also tried a variety of alternative design choices. When we use additive interaction in gate units, the result is **66.76%** mean IoU on the PASCAL validation set. In comparison, our proposed method with an element-wise product yields **68.7%** mean IoU on PASCAL val set. Intuitively, a multiplicative mechanism allows for strong modulation of representations deemed to be incorrect by higher layer features. In the additive case, while activation may be boosted for the correct representations by higher layer features, there remains a residue of incorrect representations that may weaken final predictions. Multiplicative gating is demonstrably valid from a performance standpoint, but also intuitive in that it provides a stronger capacity to resolve ambiguity present among earlier layers.

## VII. CONCLUSION

We have presented a novel end-to-end deep learning framework for dense image labeling deemed a coarse-to-fine gated feedback refinement network. Our model produces segmentation labels in a coarse-to-fine manner. The segmentation labels at coarse levels are used to progressively refine the labeling produced at finer levels. We introduce multiple loss functions in the network to provide deep supervision at multiple stages of the network. We also introduce gate units that effectively modulate signals passed forward from encoder to decoder in order to resolve ambiguity. Experimental results on several challenging dense labeling datasets demonstrate that the proposed model performs either comparable to, or significantly better than state-of-the-art approaches. Although we focus on semantic segmentation, the proposed architecture is quite general and can be applied to other pixel-wise labeling problems. In addition, experimental results based on ablation analysis reveal generality in the value of coarse-to-fine predictions, deep supervision, skip connections, and gated refinement with the implication that a wide array of canonical neural network architectures may benefit from these simple architectural modifications.

Many interesting research questions arise from the approach and results presented in this paper. One especially fruitful avenue for further investigation is to remove focus on the correctness of predictions possible at intermediate stages given that skip connections and gating allow for repair or modulation of erroneous representations. In practice, this suggests the possibility of error correcting iterative gated refinement as an interesting and important direction for future work, which may also allow for networks with a stronger representational capacity made possible by the additional slack afforded by ambiguity resolving mechanisms proposed in G-FRNet.